\documentclass[10pt,twocolumn,letterpaper]{article}
\usepackage{url}
\usepackage{xcolor}
\definecolor{newcolor}{rgb}{.8,.349,.1}

\usepackage{hyperref}
\hypersetup{
 colorlinks}

\usepackage{times}
\usepackage{epsfig}
\usepackage{graphicx}
\usepackage{amsmath}
\usepackage{amssymb}
\usepackage{amsfonts}
\usepackage{booktabs}
\usepackage{caption}
\usepackage{makecell}
\usepackage{color}
\usepackage{subfigure}

\usepackage[switch,pagewise]{lineno} 

\usepackage{wacv}
\usepackage{times}
\usepackage{epsfig}
\usepackage{graphicx}
\usepackage{amsmath}
\usepackage{amssymb}
\usepackage{amsfonts}
\usepackage{booktabs}
\usepackage{caption}
\usepackage{makecell}
\usepackage{color}
\usepackage{subfigure}

\wacvfinalcopy 

\def\wacvPaperID{000} 

\ifwacvfinal\fi
\begin{document}

\title{DAR-Net: Dynamic Aggregation Network for Semantic Scene Segmentation}

\author{Zongyue Zhao \hspace{2cm} Min Liu \hspace{2cm} Karthik Ramani\\
Purdue University\\
{\tt\small zhao938@purdue.edu}
}

\maketitle
\ifwacvfinal\thispagestyle{empty}\fi

\begin{abstract}
Traditional grid or neighbor-based static pooling has become a constraint for point cloud geometry analysis. In this paper, we propose DAR-Net, a novel network architecture that focuses on dynamic feature aggregation. The central idea of DAR-Net is generating a self-adaptive pooling skeleton that considers both scene complexity and local geometry features. Providing variable semi-local receptive fields and weights, the skeleton serves as a bridge that connect local convolutional feature extractors and a global recurrent feature integrator. Experimental results on indoor scene datasets show advantages of the proposed approach compared to state-of-the-art architectures that adopt static pooling methods.
\end{abstract}

\section{Introduction}
\label{sec1}
For the task of 3D geometry understanding, neural networks that directly take point clouds as input have shown advantages compared to voxel and multi-view based networks  that simulate 2D scenarios~\cite{maturana_voxnet:_2015,charles_pointnet:_2017,dai_scannet:_2017,wang_voting_2015,tchapmi_segcloud:_2017,qi_volumetric_2016}. The trailblazer, PointNet, addressed the lack of correspondence graph by using multi-layer perceptron and a single global pooling layer, neither of which relied on local dependencies~\cite{charles_pointnet:_2017}. The two-scale network performed well on object analysis. However, its deficiencies in a) local correspondence identification; b) intermedium feature aggregation and c) valid global information integration lead to poor performance on large-scale scene segmentation.

\begin{figure}
 \centering
 \includegraphics[width=0.47\textwidth]{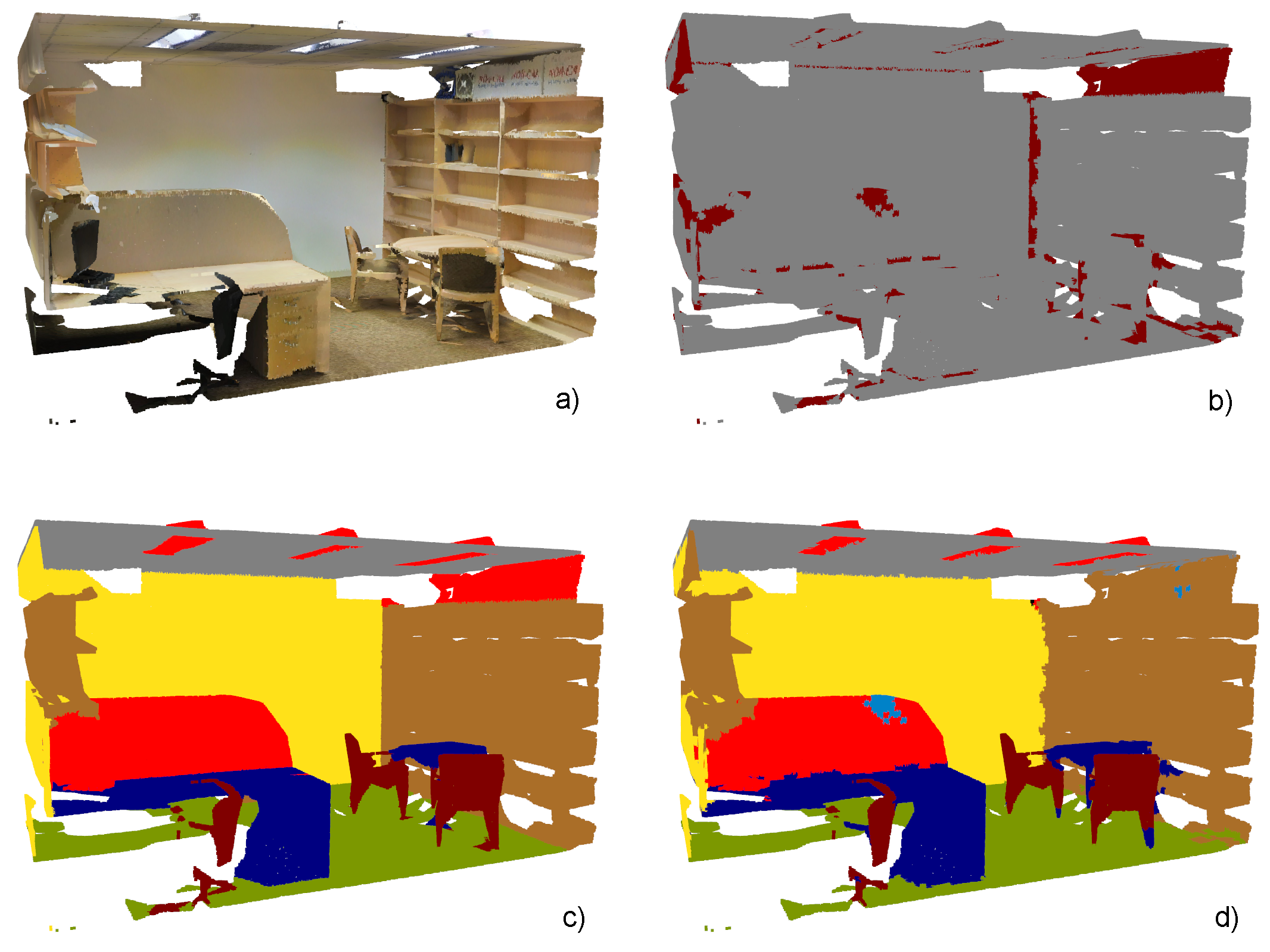}
 \caption{Segmentation results on the S3DIS dataset. a) Input point cloud; b) Validation: color grey indicates successful prediction; c) Ground truth; d) Prediction from DAR-Net.}
 \label{fig:visual}
\end{figure}

Analyzing the drawbacks of PointNet, several papers worked on the local deficiency by constructing mapping indices for convolutional neural networks (CNN)~\cite{tatarchenko_tangent_2018,boscaini_learning_2016,li_pointcnn:_2018,shoef_pointwise:_2019}. On the other hand, works that focused on the global integration problem gained inspiration from natural language processing and turned to deep recurrent neural networks (RNN)~\cite{tchuinkou_r-covnet:_2018,huang_recurrent_2018,ferrari_3d_2018}. While various works contributed to both micro and macro ends of the scale spectrum, what left in between was less attended. Feature aggregation between local neighborhoods and global representation, if any, remain to be static and independent of the geometry context~\cite{boscaini_learning_2016,huang_recurrent_2018,liu_3dcnn-dqn-rnn:_2017,qi_pointnet++:_2017,tchuinkou_r-covnet:_2018,tatarchenko_tangent_2018,ferrari_3d_2018}. For example, Tangent Convolutions~\cite{tatarchenko_tangent_2018} used rectangular grids with uniform resolution for local mean pooling. RSNet~\cite{huang_recurrent_2018} evenly divided the entire scene into slices and did max-pooling within each slice. 3P-RNN~\cite{ferrari_3d_2018}, despite introducing variance of receptive field sizes in the local scale, went back to the voxelization track when feeding the global recurrent network. Those rigid pooling methods adapted no information density distribution within the point cloud, leading to computational inefficiencies and poor segmentation results on less-dominating classes. Shapes with rich geometry features but occurred less are not detected effectively.

We present an approach for intermedium feature aggregation to address deficiencies from traditional static pooling layers. The key concept in the aggregation process is forming a pooling skeleton whose a) size is corresponding to the individual scene scale and complexity; b) each node links a variable set of points that represent a meaningful spatial discreteness; c) each node is weighted against the index set to further utilize information distribution, and provide robustness even when the node-discreetness correlation fails. Such a skeleton is unsupervised learned prior to the training process.

We construct a network, DAR-Net, to incorporate the dynamic aggregation operation with convolutional feature extraction and global integration, while handling permutation invariance in multiple scales. The network is trained on two large-scale indoor scene datasets~\cite{dai_scannet:_2017,armeni_3d_2016} and it shows advantage compared to recent architectures using static pooling methods and similar inputs. A sample of the semantic segmentation results on the S3DIS dataset~\cite{armeni_3d_2016} (Sec. \ref{data}) is shown in Figure \ref{fig:visual}.

\section{Related Work}
Recent contributions relevant to our work can be roughly divided into three categories: convolutional feature extraction, global integration and unsupervised pre-processing. For context completeness, traditional 3D analysis networks that do not operate on point clouds are first introduced. 

\subsection{Prior to point clouds}
Although convolutional neural networks (CNN) had achieved great success in analyzing 2D images, they cannot be directly applied to point clouds beacuse of its unorganized nature. Without a pixel-based neighborhood defined, vanilla CNNs cannot extract local information and gradually expand receptive field sizes in a meaningful manner. Thus, segmentation tasks were first performed in a way that simulate 2D scenarios – by fusing partial views represented with RGB-D images together~\cite{afzal2014rgb,qi_volumetric_2016,lun20173d,hazirbas_fusenet:_2016}. Some other work transform point clouds into cost-inefficient voxel representations on which CNN can be directly applied~\cite{maturana_voxnet:_2015,huang_point_2016,dai_scannet:_2017}. 

While these methods did benefit from mature 2D image processing network structures, inefficient 3D data representations constrained them from showing good performance for scene segmentation, where it is necessary to deal with large, dense 3D scenes as a whole. Therefore, recent research gradually turned to networks that directly operate on point clouds when dealing with semantic segmentation for complex indoor/outdoor scenes~\cite{ferrari_3d_2018,tatarchenko_tangent_2018,landrieu_large-scale_2018}. 

\subsection{Local feature extraction}
As introduced, PointNet used multi-layer perception (which process each point independently) to fit the unordered nature of point clouds~\cite{huang_point_2016}. Furthermore, similar approaches using $1\times1$ convolutional kernels~\cite{li_so-net:_2018}, radius querying~\cite{qi_pointnet++:_2017} or nearest neighbor searching~\cite{klokov_escape_2017} were also adopted. Because local dependencies were not effectively modeled, overfitting constantly occurred when these networks were used to perform large-scale scene segmentation. In addition, work like R-Conv~\cite{tchuinkou_r-covnet:_2018} tried to avoid time-consuming neighbor searching with global recurrent transformation prior to convolutional analysis. However, scalability problems still occurred as the global RNN cannot directly operate on the point cloud representing an entire dense scene, which often contains millions of points.

Tangent Convolution \cite{tatarchenko_tangent_2018} proposed a way to efficiently model local dependencies and align convolutional filters on different scales. Their work is based on local covariance analysis and down-sampled neighborhood reconstruction with raw data points. Despite tangential convolution itself functioned well extracting local features, their network architecture was limited by static, uniform medium level feature aggregation and a complete lack of global integration.

\subsection{Global Integration}
Several works turned to the global scale for permutation robustness. Its simplest form, global maximum pooling, only fulfilled light-weight tasks like object classification or part segmentation~\cite{charles_pointnet:_2017}. Moreover, RNNs constructed with advance cells like Long-Short-Term-Memory~\cite{hochreiter_long_1997} or Gate-Recurrent-Unit~\cite{chung_empirical_2014} offered promising results on scene segmentation~\cite{landrieu_large-scale_2018}, even for those architectures without significant consideration for local feature extraction~\cite{huang_recurrent_2018,liu_3dcnn-dqn-rnn:_2017}. However, in those cases the global RNNs were built deep, bidirectional or compact with hidden units, giving out a strict limitation on the direct input. As a result, the original point cloud was often down-sampled to an extreme extent, or the network was only capable of operating on sections of the original point cloud~\cite{huang_recurrent_2018}.

\subsection{Unsupervised Learning}
Various works in this area aimed to promote existing supervised-learning networks as auto-encoders. For example, FoldingNet~\cite{yang_foldingnet:_2018} managed to learn global features of a 3D object through constructing a deformable 2D grid surface; PointWise~\cite{shoef_pointwise:_2019} considered theoretical smoothness of object surface; and, MortonNet~\cite{thabet_mortonnet:_2019} learned compact local features by generating fractal space-filling curves and predicting its endpoint. Although features provided by these auto-encoders are reported to be beneficial, we do not adopt them into our network for a fair evaluation on the aggregation method we propose.

Different from the common usage of finding a rich, concise feature embedding, SO-Net~\cite{li_so-net:_2018} learned a self-organizing map (SOM) for feature extraction and aggregation. Despite its novelty, few performance improvements were observed even when compared to PointNet or OctNet~\cite{riegler_octnet:_2017}.
Possible reasons include the lack of a deep local and global analysis. 

SO-Net used the SOM to expand the scale of data for local feature extraction, and conducted most of the operations on the expanded feature space. The architecture was capable of handling object analysis tasks. However, for this task each point cloud merely contained a few thousand points, making the benefit from carefully arranging tens of pooling nodes limited. We argue that SOM or other similar self-adapted maps perform better when used to contract the feature space for analyzing large-scale point clouds. Map nodes should be assigned with appropriate weights to provide more detailed differentiation and robustness. Once features are aggregated to the skeleton nodes, a thorough, deep integration process should be conducted prior to decoding.

\section{Dynamic aggregation}

This section aims to transform pointwise local features to semi-local rich representations, and, to propagate information stored in the semi-local space back to the point cloud for semantic analysis. In this process, two things need to be properly designed. First, a pooling skeleton (intermedium information carrier) that adapts global and local geometry structures of the input point cloud. Second, reflections that map the skeleton feature space from and to the point cloud feature space. 

For clarity, in following paragraphs the point cloud is referred as $P_N=\{p_i\mid 0<i\leq N\}$ and its pooling skeleton is referred as $S_M=\{s_j\mid 0< j\leq M\}$. 

\subsection{Skeleton formation} \label{form}

For the task of indoor scene segmentation, the scale of each individual scene varies significantly. E.g., in S3DIS dataset~\cite{armeni_3d_2016} the most complicated scene contains $9.2\times 10^6$ points, more than 100 times larger than the least complicated one ($8.5\times 10^4$). Therefore, the size of the skeleton, indicated with the number of nodes $M$, should not be static like those work applied to object analysis \cite{li_so-net:_2018}. Further ablation studies (Sec. \ref{ABl}) demonstrate that an empirical logarithm relationship (Figure \ref{fig:rec}) between $N$ and $M$ better adapts scene complexity than stationary $M$ or stationary average receptive field size.

We use a Kohonen network~\cite{kohonen_self-organized_1982,li_so-net:_2018} to implement the dynamic skeleton. Contrary to initialization methods suggested by~\cite{ciampi_clustering_2000,li_so-net:_2018}, we conduct random initialization prior to normalization to provide a reasonable guess with respect to substantial spacing along different axes. Such a method provides extra robustness for the task of scene understanding, where individual inputs often contain a dimension disproportionate to others (long hallways, large conference rooms). 

\begin{figure}
 \centering
 \includegraphics[width=0.4\textwidth]{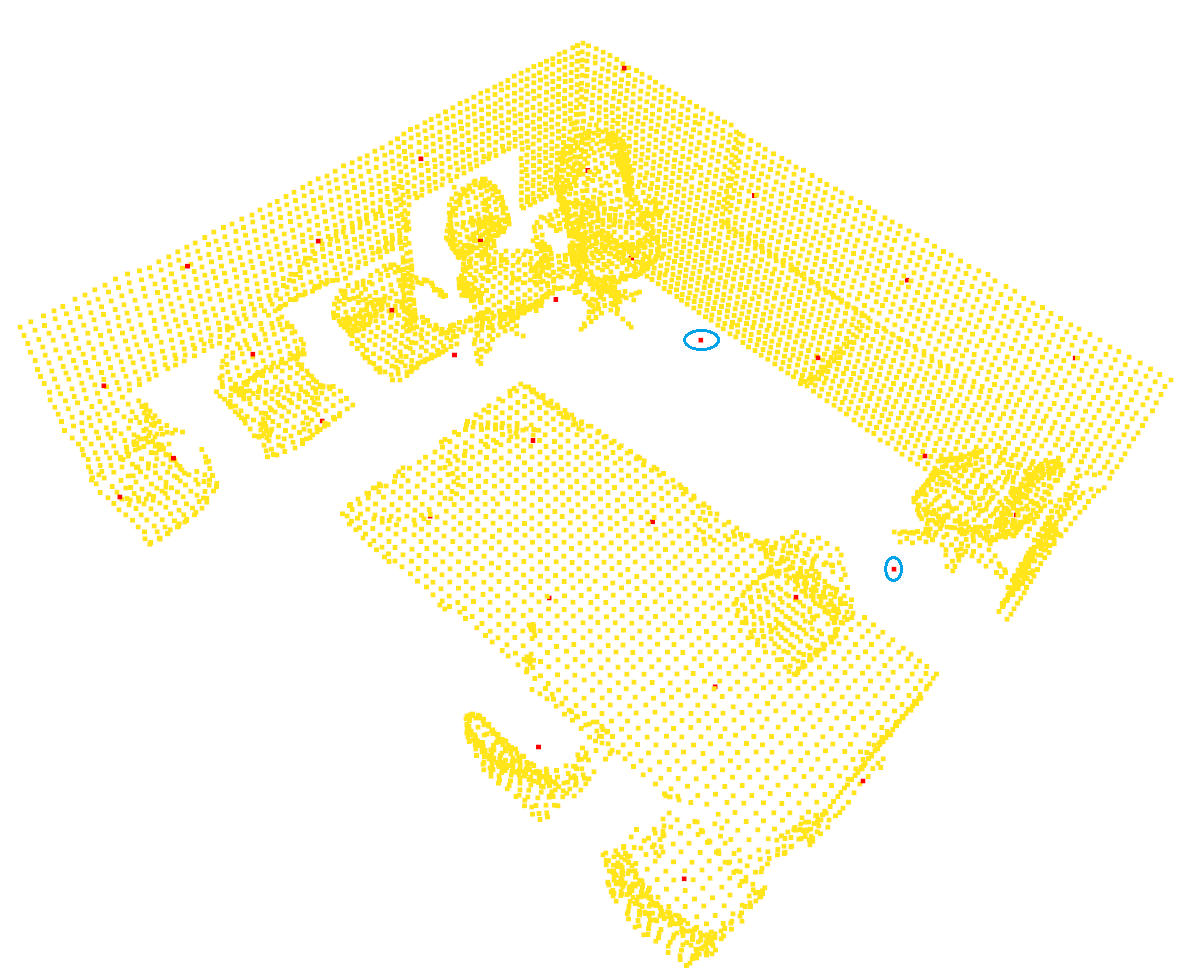}
 \caption{Dynamic Aggregation Skeleton (Red) and Point Cloud (Yellow). (Floor, ceiling and walls in the front are removed for clarity.) Note that each chair is assigned with at least one skeleton node, while uniform structures like walls are assigned with smaller node density. Two nodes (circumscribed with blue ovals) fail to be attached to point cloud. Such a problem is addressed in Sec. \ref{Agg}.}
 \label{fig:som}
\end{figure}

An example of the skeleton is shown in Figure \ref{fig:som}.

\subsection{Feature aggregation} \label{Agg}

Consider an arbitrary point cloud $P_N$ and its corresponding skeleton $S_M$, dynamic aggregation maps pointwise feature space $F^{agg-i}\subset \Re^{N\times C}$ into the node-wise feature space $F^{agg-o}\subset \Re^{M\times C}$. By introducing a correspondence indicator $T_j:0<T_j\leq N$ that regulate node receptive field size and a possible global intervention factor $g$, the general expression of dynamic aggregation is shown in Eq. (\ref{general}).
\begin{equation} \label{general}
f_j^{agg-o}=f_j^{agg-o}(f_{i^j_1}^{agg-i},...,f_{i^j_{T_j}}^{agg-i},g),~0< i^j_t\leq N
\end{equation}

Such general expression contains two sections that await instantiation: the dependency searching function that constructs indices $\{i_t\}$, and the pooling function dealing with an arbitrary set of inputs $f=f^{agg-o}(f_1,...,f_T,g)$. 

\textbf{Indexing.} Each point $p_i$ is first linked to its K-nearest neighbor nodes, referred as $\{s_{iK}\}$. As all points are iterated through this process, a global index matrix $I\subset \mathbb{N}^{N\times K}$ is formed, whose element $I(i,k)$ is the index of the k-th neighbor node point $p_i$ searched, i.e., $\forall ind \in I, 0< ind\leq M $.  Dynamic aggregation indices, $\{i_t\}$, are then generated from traversing the node space: all points indexing $s_j$ will be categorized into $\{i_t^j\}=\{i\mid I(i,k)=s_j\}$.

\begin{figure*}
 \centering
 \includegraphics[width=0.95\textwidth]{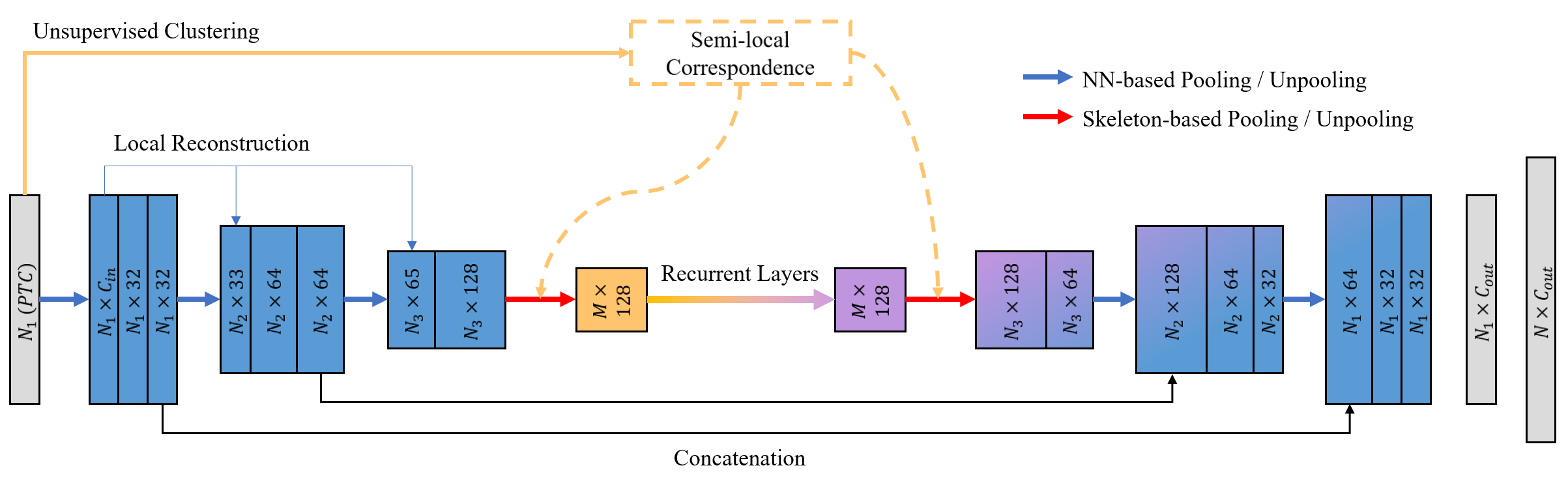}
 \caption{Schematic Diagram of DAR-Net. Solid blocks represent features: $N_i$ indicate pointwise features on $\{p_i\}$; $M$ indicate node-wise features on $\{s_j\}$. Color information is also used as input since geometrical structures are not sufficient for identifying multiple classes (Sec. \ref{data}).}
 \label{fig:arc}
\end{figure*}

\textbf{Aggregating function.} A semi-average pooling method is used to further extract information density and address skeleton construction failures. Although skeleton nodes are already arranged more compact where geometry features vary more (Sec. \ref{form}), each node in those areas is still indexing relevantly more neighbor points, as rich geometry features usually fold more points in unit space (in the scale of skeleton receptive field). Therefore, assigning a larger weight to nodes correlating with more points becomes advantageous. Moreover, when a skeleton node fails to represent any geometry structure (see examples shown in Figure \ref{fig:som}), traditional average or maximum pooling cannot identify the situation and passes irrelevant features forward. 

The semi-average pooling function, shown in Eq. (\ref{pooling}), is implemented with a global descriptor $g$ that indicates average reception field size, i.e., the average amount of neighbor points a node would index.

\begin{equation} \label{pooling}
\left\{
\begin{array}{lr}
f^{agg-o}_j=\sum_j(f_{i^j_1}^{agg-i},...,f_{i^j_{T_j}}^{agg-i})/g & \\
\\
g=\sum_j|\{i_t^j\}|/M=\sum_jT_j/M & 
\end{array}
\right.
\end{equation}

\subsection{Feature propagation}

As the neighborhood searching process is conducted on the node space, the global index matrix $I$ can be directly used to unpool node-wise features back to the point cloud as if pooling from $\Re^{KN\times C}$ to $\Re^{N\times C}$.
\begin{equation} \label{unpool}
f^{ppg-o}_i=\max_k\{f^{ppg-i}_{I(i,k)}\}
\end{equation}

Where $f^{ppg-o}_i$ is the propagated feature corresponding to point $p_i$ and $f^{ppg-i}_{I(i,k)}$ are features corresponding to nodes indexed by point $p_i$.
Note that the redundant space (size of $KN$) is only implicitly used with indices throughout the pooling-unpooling process. Hence, the dynamic aggregation approach we propose is compatible with large-scale dense point clouds.

\section{Global integration}
Global Integration aims to model the long-range dependency in a point cloud, which can be described as a reflection mapping one node-wise feature space to another: $R:\Re^{M\times C_1}\rightarrow \Re^{M\times C_2}$. 

We use GRU-based RNN~\cite{chung_empirical_2014} for permutation invariance upon unordered nodes. Features on the entire skeleton, $\{f^{agg-o}\}\subset \Re^{M\times C}$, are treated as a single-batch sequence $Q$ of length $M$: $Q[j]=f^{agg-o}_j$. In addition, as $M$ varies from scene to scene, all input sequences are padded to the same length $\max \{M\}$. The padded sequence is then fed into the recurrent network. As a result, output features on each node is relevant to input information from all nodes, creating a maximized overlapping receptive field:
\begin{equation}
 f^{rnn-o}_j=R(f^{agg-o}_1,...,f^{agg-o}_M)
\end{equation}

\section{Architecture}

We design a convolutional-recurrent network to consort short and long range spatial dependencies with the operation of dynamic aggregation, as is shown in Figure \ref{fig:arc}. The skeleton-based aggregation in this network clusters intermediate level information and compresses the feature space. Thus, the recurrent integration network is designed compact and efficient.

For pre-processing, we first estimate the scene complexity to determine the appropriate skeleton size for each scene. The skeleton is then unsupervised clustered for a preliminary understanding of the semi-local structure distribution. 

We adopt tangent convolutions~\cite{tatarchenko_tangent_2018} for local pointwise feature extraction. The encoded local features are then dynamically aggregated to the skeleton, as an intermediate scale of information abstraction. Furthermore, node-wise features that independently correspond to an intermediate receptive field are treated with a global RNN, which implicitly learns long-range knowledge. Globally integrated information is propagated back to the point cloud for concatenations with local features and hierarchically decoding. In the end, $1\times1$ pointwise convolutions are used to generate semantic prediction results.

\section{Experiments}

In this section, we first report a few implementation details and evaluation criteria, then present best segmentation results DAR-Net generates. More experiments are discussed in the ablation study section. 

\begin{table*} 
\centering
\scalebox{0.81}{
\begin{tabular}{l|c|c|ccccccccccccc}
\toprule
Method & \textbf{mIoU} & \textbf{mA} & ceiling & floor & wall & beam & column & window & door & chair & table & sofa & bookcase & board & clutter\\
\midrule
PointNet~\cite{charles_pointnet:_2017} & 41.1 & 49.0 & 88.8 & 97.3 & 69.8 & {0.1} & 3.9 & {49.3} & 10.8 & 58.9 & 52.6 & 5.8 & 40.3 & 26.3 & 33.2 \\
SEGCloud~\cite{tchapmi_segcloud:_2017} & 48.9 & 57.4 & 90.1 & 96.1 & 69.9 & 0.0 & 18.4 & 38.4 & 23.1 & {75.9} & 70.4 & 58.4 & 40.9 & 13.0 & 41.6 \\
T-Conv~\cite{tatarchenko_tangent_2018} & 52.8 & 62.2 & - & - & - & - & - & - & - & - & - & - & - & - & - \\
RSNet~\cite{huang_recurrent_2018} & 51.9 & 59.4 & 93.3 & {98.3} & {79.2} & 0.0 & 15.8 & 45.4 & 50.1 & 65.5 & 67.9 & 52.5 & 22.5 & 41.0 & 43.6 \\
Ours & {58.8} & {68.4} & {93.4} & 97.0 & 75.3 & 0.0 & {26.6} & 47.2 & {50.8} & 68.2 & {81.3} & {61.7} & {63.4} & {53.7} & {46.0} \\
\bottomrule
\end{tabular}}
\caption{\textbf{Results on the S3DIS Dataset}}
\label{table1}
\end{table*}

\subsection{Datasets and details} \label{data}

The performance of dynamic segmentation and DAR-Net is evaluated in the task of indoor scene segmentation. Two commonly used large-scale datasets, Stanford Large-Scale 3D Indoor Spaces (S3DIS)~\cite{armeni_3d_2016} and ScanNet~\cite{dai_scannet:_2017} are adopted to conduct the experiments.

\textbf{S3DIS} dataset includes more than 200 dense indoor scenes gathered from three different buildings, each scene contains up to more than nine million points in 13 classes. For this dataset, we use a widely-used A5 train-test split ~\cite{ferrari_3d_2018,huang_recurrent_2018,tchapmi_segcloud:_2017,charles_pointnet:_2017}.

\textbf{ScanNet} dataset contains over 1,500 indoor scans labeled in 20 classes. We use the standard train-test spilt provided by \cite{dai_scannet:_2017}.

\textbf{Implementation details.} We introduce multiple levels of feature carriers other than the most compact $M$ space. For clarity, point clouds down-sampled to a certain resolution $r$ will be denoted as $P^r$. 

For computational purposes, we use coordinates and color information on $P^5$ as raw inputs. Coordinates are then used to generate the dynamic pooling skeleton and conduct covariance analysis, for estimating normals and reconstructing local neighborhoods. As a result, input channels of the feature extractor include depth to the tangential plane, z-coordinate, estimated normals and RGB information, all of which are batch-normalized to $[0,1]$~\cite{tatarchenko_tangent_2018}. Feature extractors encode $p^5$ to a rich representation on $P^{20}$ with 128 channels~\cite{tatarchenko_tangent_2018}. Features on $P^{20}$ are then aggregated to the skeleton space $S^M$, which is concise enough for a global integration network to handle ($M\leq 256$). 

For best performance (Sec. \ref{ABl}), the RNN is designed to be single-directional, single-layered with 256 hidden units. Its 128 output channels are then propagated to $P^{20}$ and fed back to convolutional decoders.

All our reported results are based on original point clouds. As the network only gives out segmentation on the down-sampled $P^5$ space, a nearest neighbor searching between $P^5$ and $P$ is conducted to extrapolate predictions.

The only data augmentation method we adopt is rotation about the z-axis, in order to reduce invalid information from normal vector directions.

We use individual rooms as training batches. Following the suggestions of~\cite{tatarchenko_tangent_2018}, we pad each room to an uniform batch size throughout the network for computational purposes. Padded data stays out of indexing and has no effect.

All supervised sections are trained as a whole using the cross-entropy loss function and an Adam optimizer with an initial learning rate of $5\times10^{-4}$~\cite{kingma2014adam}.

The unsupervised clustering network is trained separately. The Kohonen map is initialized for each individual scene to accommodate its shape and complexity. Qualitative results suggest that the usage of prime components, either as additional inputs or for initial guesses, hurts the robustness of the algorithm. Therefore, the Kohonen maps are initialized randomly with respect to the nodes spacing, and trained to convergence with the initial learning rate of 0.4.

\textbf{Measures.} For quantitative reasoning, we report mean (over classes) intersection over union (\textbf{mIoU}), class-wise \textbf{IoU} and mean accuracy over class (\textbf{mA}). We do not use the indicator of overall accuracy as it fails to measure the actual performance for scene segmentation, where several classes (floor, ceiling, etc.) are dominating in size yet easy to identify. In addition, all results are calculated over the entire dataset, i.e., if a certain class does not occur in a certain scene, we do NOT pad accuracy for misleadingly better results. 

\begin{table*}[t]
\centering
\scalebox{0.80}{
\begin{tabular}{l|ccccccccccccccc}
\toprule
Method& \multicolumn{1}{c|}{\textbf{mIoU}} & \multicolumn{1}{c|}{\textbf{mA}}& Wall & Floor& Chair & Table  & Desk& Bed & Shelf & Sofa & Sink\\
\midrule
PointNet~\cite{charles_pointnet:_2017}& \multicolumn{1}{c|}{14.7} & \multicolumn{1}{c|}{19.9} & 69.4 & 88.6 & 35.9  & 32.8& 2.6 & 18.0& 3.2  & 32.8 & 0 \\
PointNet++~\cite{qi_pointnet++:_2017} & \multicolumn{1}{c|}{34.3} & \multicolumn{1}{c|}{43.8} & 77.5 & 92.5 & 64.6  & 46.6& 12.7& 51.3& 52.9 & 52.3 & 30.2\\
T-Conv~\cite{tatarchenko_tangent_2018}& \multicolumn{1}{c|}{40.9} & \multicolumn{1}{c|}{55.1} & -  & -  & -& - & - & - & - & -  & - \\
RSNet~\cite{huang_recurrent_2018}& \multicolumn{1}{c|}{39.4} & \multicolumn{1}{c|}{48.4} & {79.2} & {94.1} & 65.0  & {51.0}& {34.5}& 56.0& 53.0 & 55.4 & 34.8\\
Ours  & \multicolumn{1}{c|}{41.1} & \multicolumn{1}{c|}{58.1} & 71.1&92.7&66.7&49.2&36.0&59.8&50.5&58.3&36.9 \\
\midrule
3DMV\footnotemark[1]~\cite{dai_3dmv:_2018}&\multicolumn{1}{c|}{48.4}&\multicolumn{1}{c|}{-}&60.2&79.6&60.6&41.3&43.3&53.8&64.3&50.7&47.2\\
TextureNet\footnotemark[1]~\cite{huang_texturenet:_2019} & \multicolumn{1}{c|}{56.6} & \multicolumn{1}{c|}{-} & 68.0 & 93.5 & 71.9 & 46.4 & 41.1 & 66.4 & 67.1 & 63.6 & 56.5 \\
\midrule
Method& Bathtub & Toilet& Curtain & Counter & Door & Window & Shower & Fridge & Picture& Cabinet & Others \\
\midrule
PointNet~\cite{charles_pointnet:_2017}& 0.2 & 0.0& 0.0  & 5.1& 0.0& 0.0 & 0.0 & 0.0 & 0.0 & 5.0& 0.1  \\
PointNet++~\cite{qi_pointnet++:_2017} & 42.7  & 31.4  & {33.0} & 20.0 & 2.0& 3.6 & 27.4& 18.5& 0.0 & 23.8 & 2.2 \\
T-Conv~\cite{tatarchenko_tangent_2018} & -& -& -  & -  & -& - & - & - & - & -  & - \\
RSNet~\cite{huang_recurrent_2018}& 49.4  & 54.2  & 6.8& 22.7 & 3.0& 8.8 & {29.9}& {37.9}& 1.0  & {31.3} & 19.0 \\
Ours  & 55.8& 61.6 & 24.7 & 24.8 & 18.0 & 14.5& 18.3& 30.2& 2.4 & 30.5 & 20.1\\
\midrule
3DMV\footnotemark[1]~\cite{dai_3dmv:_2018}&48.4&69.3&57.4&31.0&37.8&53.9&20.8&53.7&21.4&42.4&30.1\\
TextureNet\footnotemark[1]~\cite{huang_texturenet:_2019}& 67.2 & 79.4 & 67.8 & 44.5 & 39.6 & 56.8 & 53.5 & 41.2 & 22.5 & 49.4 & 35.6\\
\bottomrule
\end{tabular}}
\caption{\textbf{Results on the ScanNet Dataset}}
\label{table2}
\end{table*}

\footnotetext[1] {Different input channels are used (3DMV: images and voxels; TextureNet: grid points and texture patches) whereas the other networks were fed with raw points, normal vectors and color information.}

\begin{figure*}[h]
\centering
\subfigure[Color]{
\begin{minipage}[t]{0.24\linewidth}
\centering
\includegraphics[width=1\textwidth]{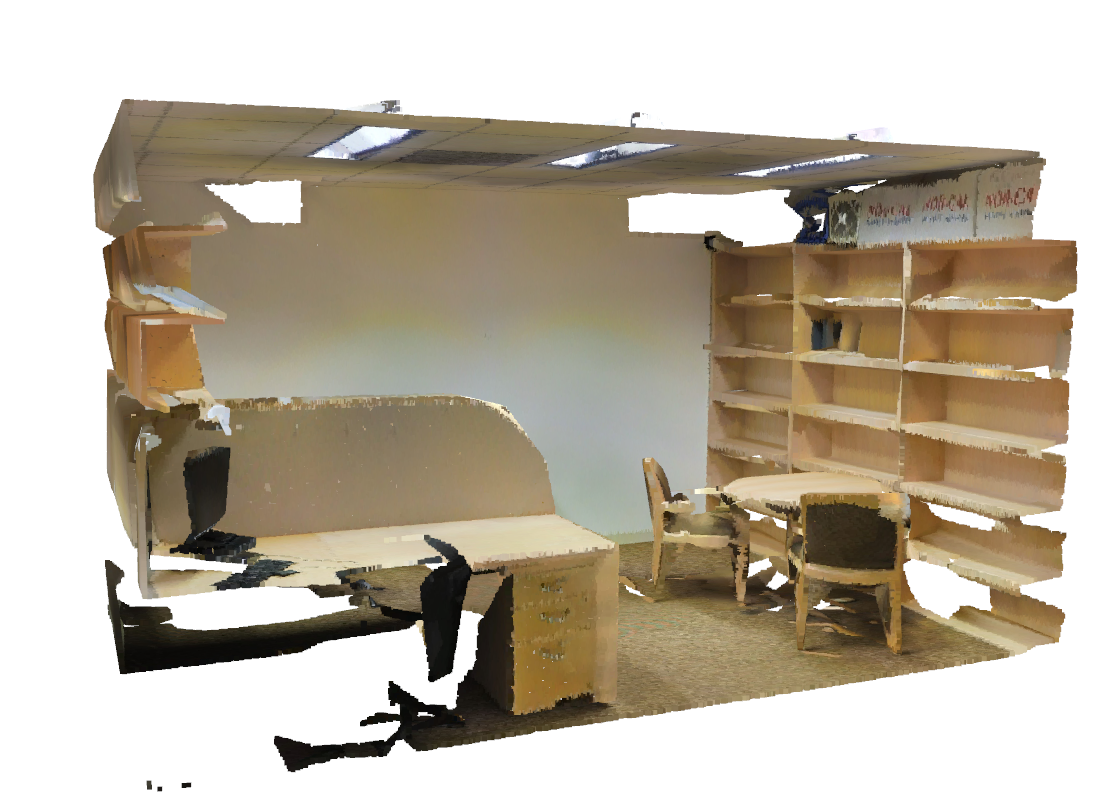}
\includegraphics[width=1\textwidth]{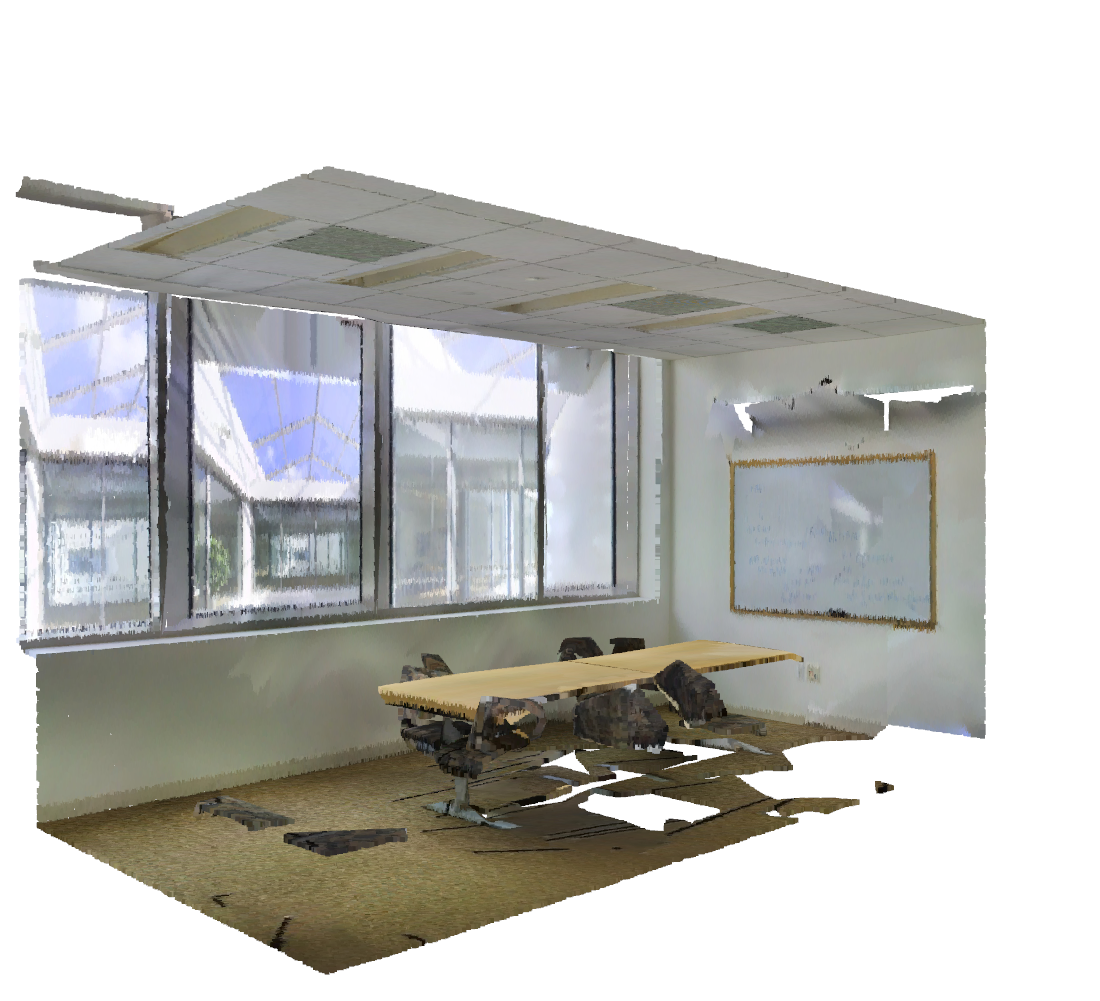}
\includegraphics[width=1\textwidth]{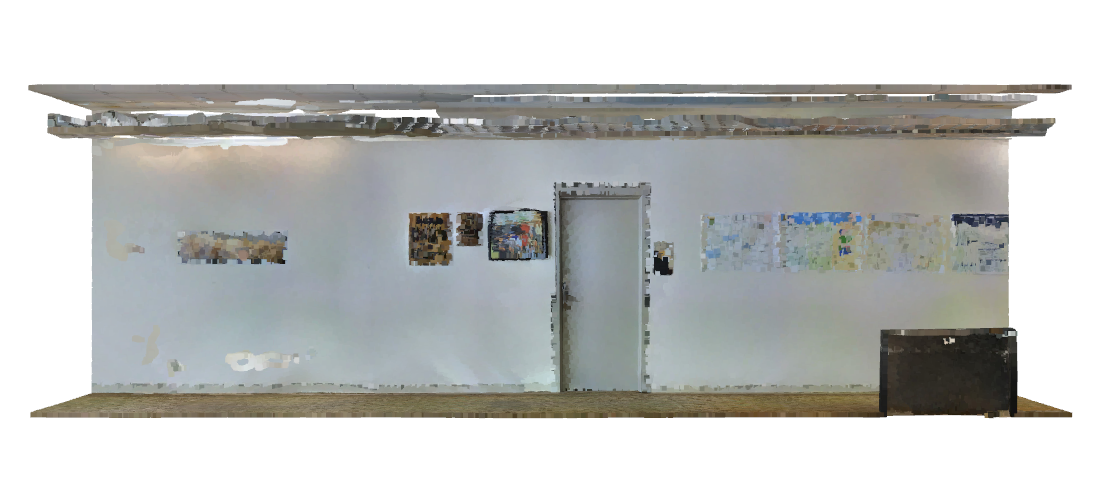}
\includegraphics[width=0.7\textwidth]{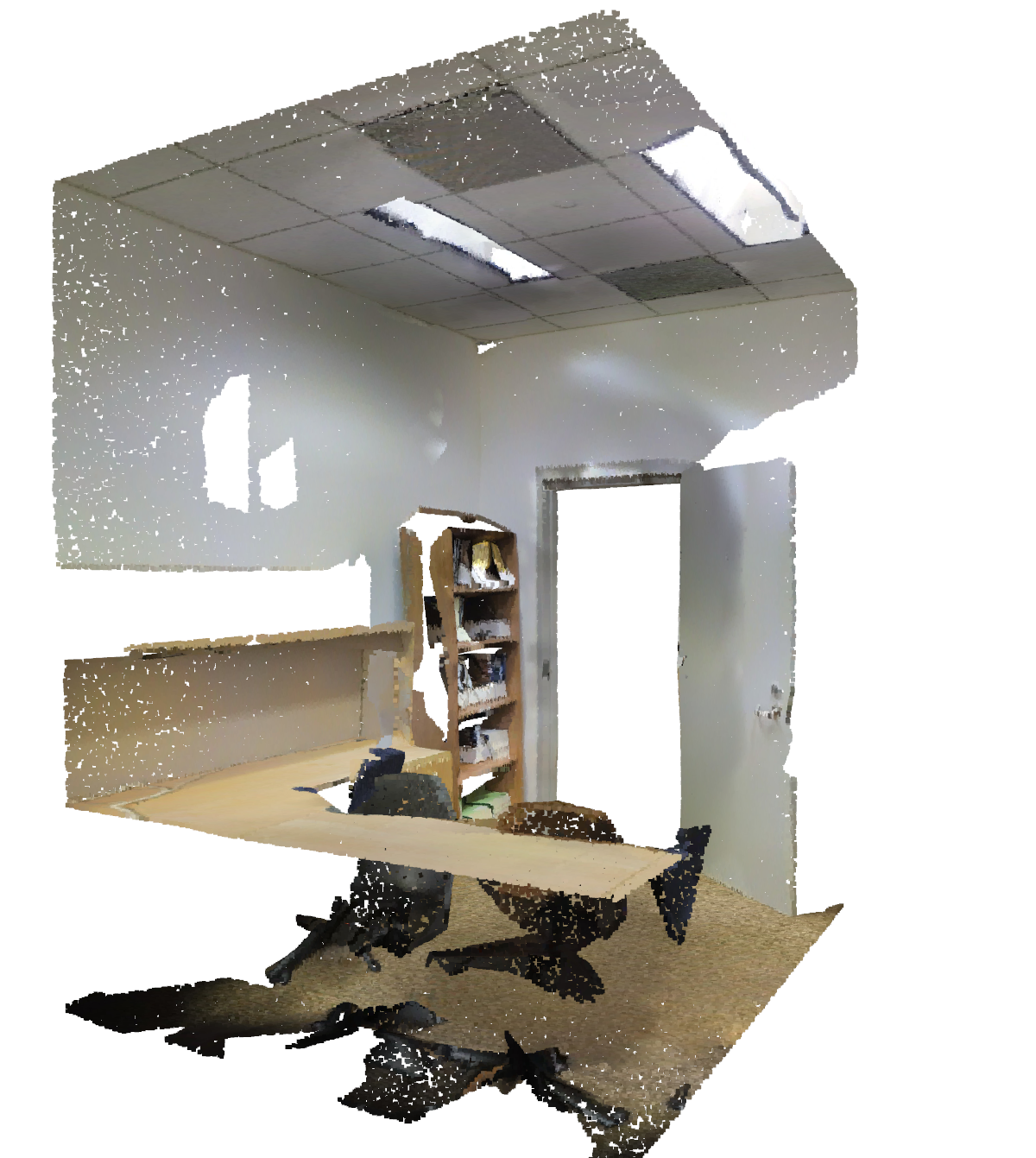}
\end{minipage}%
}%
\subfigure[Tangent Conv~\cite{tatarchenko_tangent_2018}]{
\begin{minipage}[t]{0.24\linewidth}
\centering
\includegraphics[width=1\textwidth]{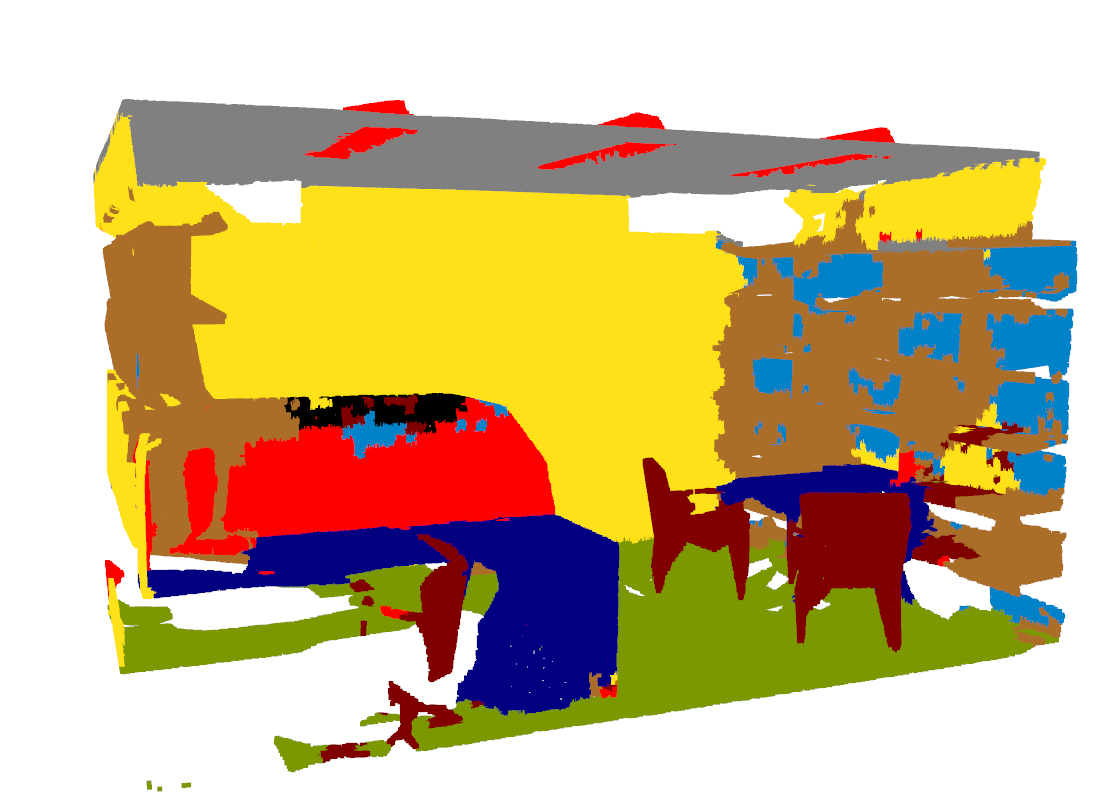}
\includegraphics[width=1\textwidth]{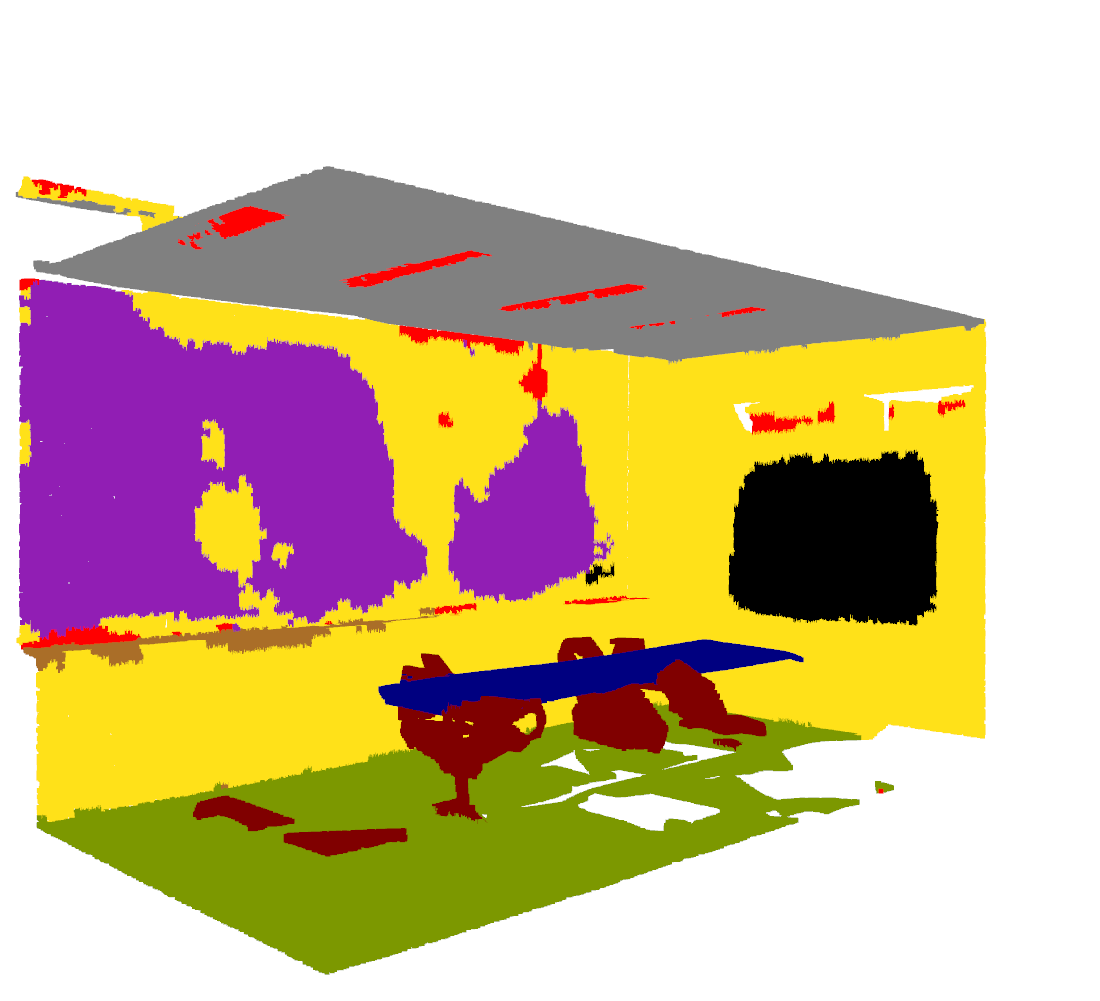}
\includegraphics[width=1\textwidth]{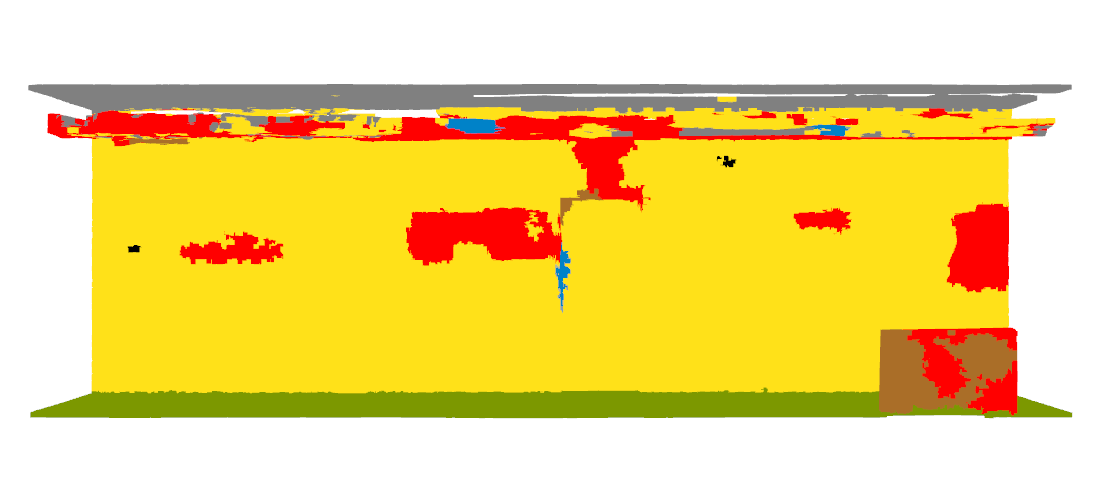}
\includegraphics[width=0.7\textwidth]{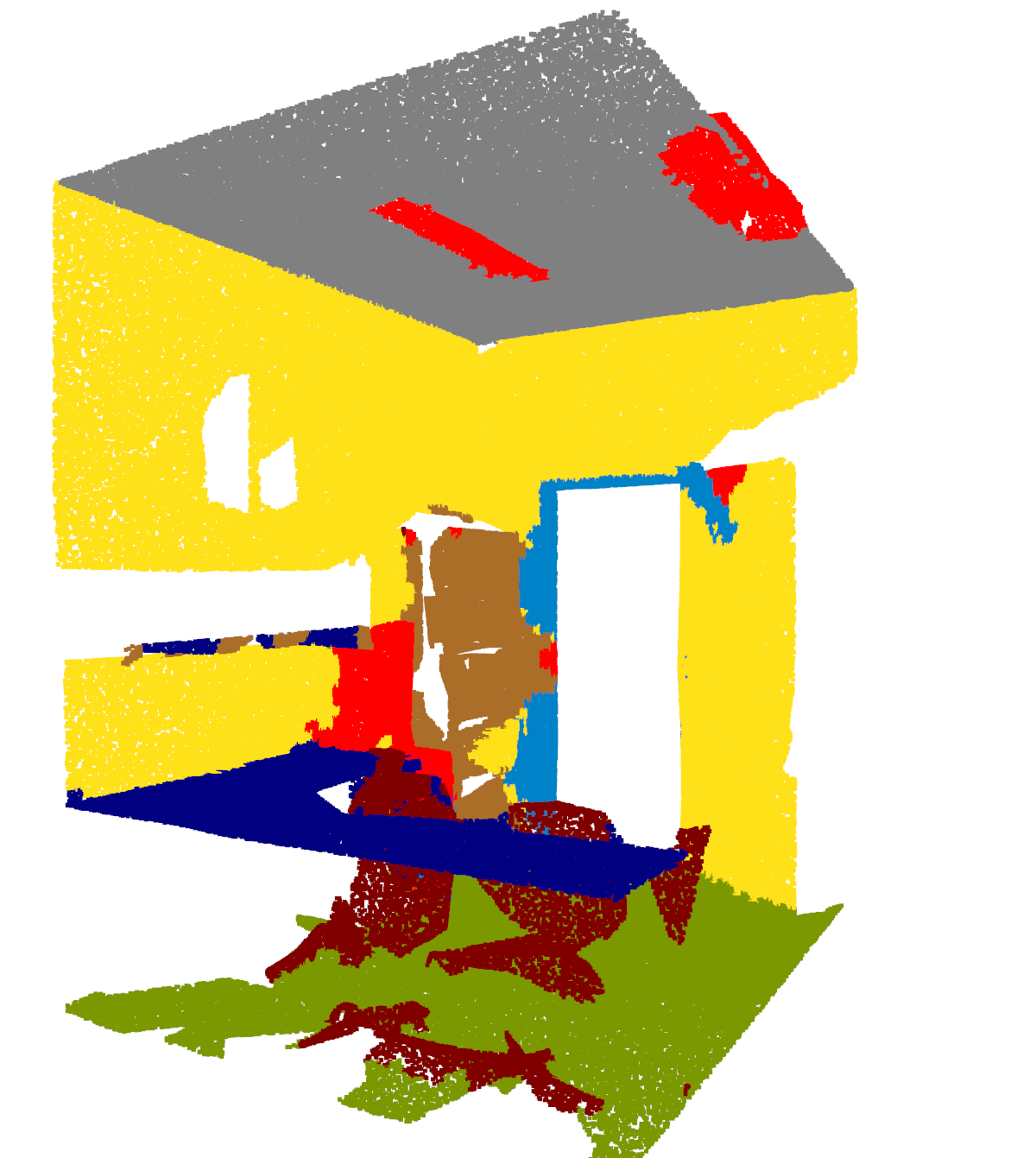}
\end{minipage}%
}%
\subfigure[Ours]{
\begin{minipage}[t]{0.24\linewidth}
\centering
\includegraphics[width=1\textwidth]{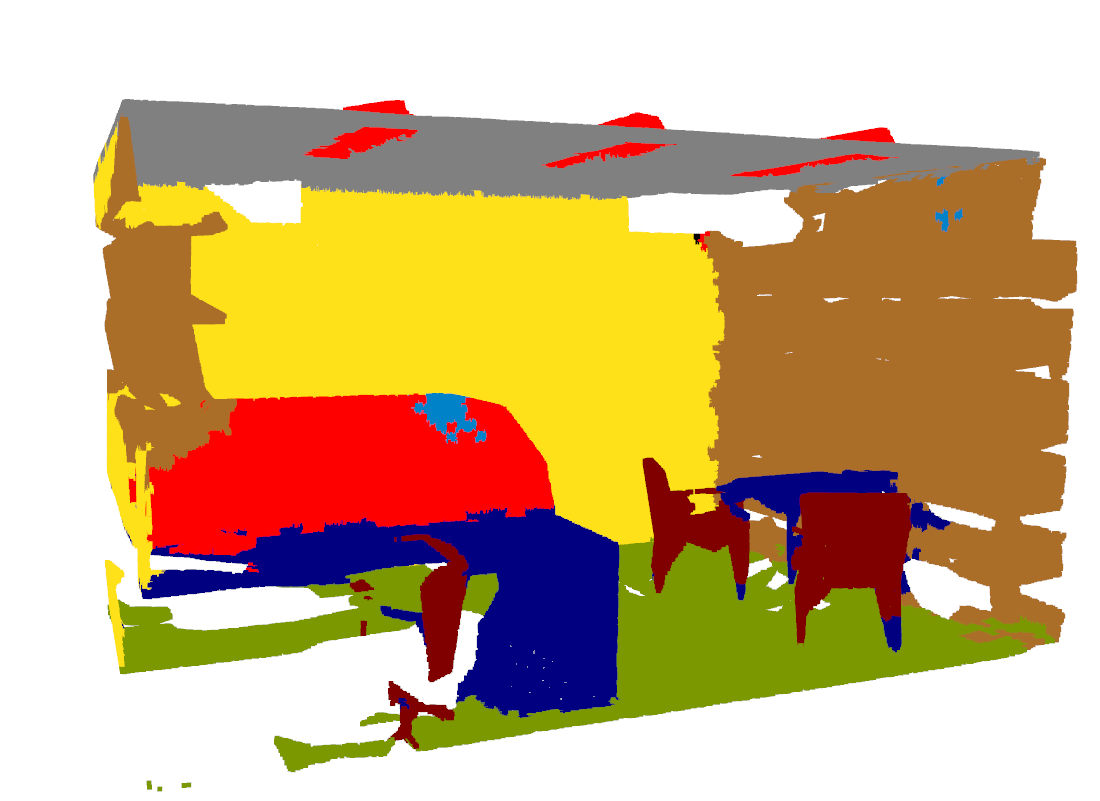}
\includegraphics[width=1\textwidth]{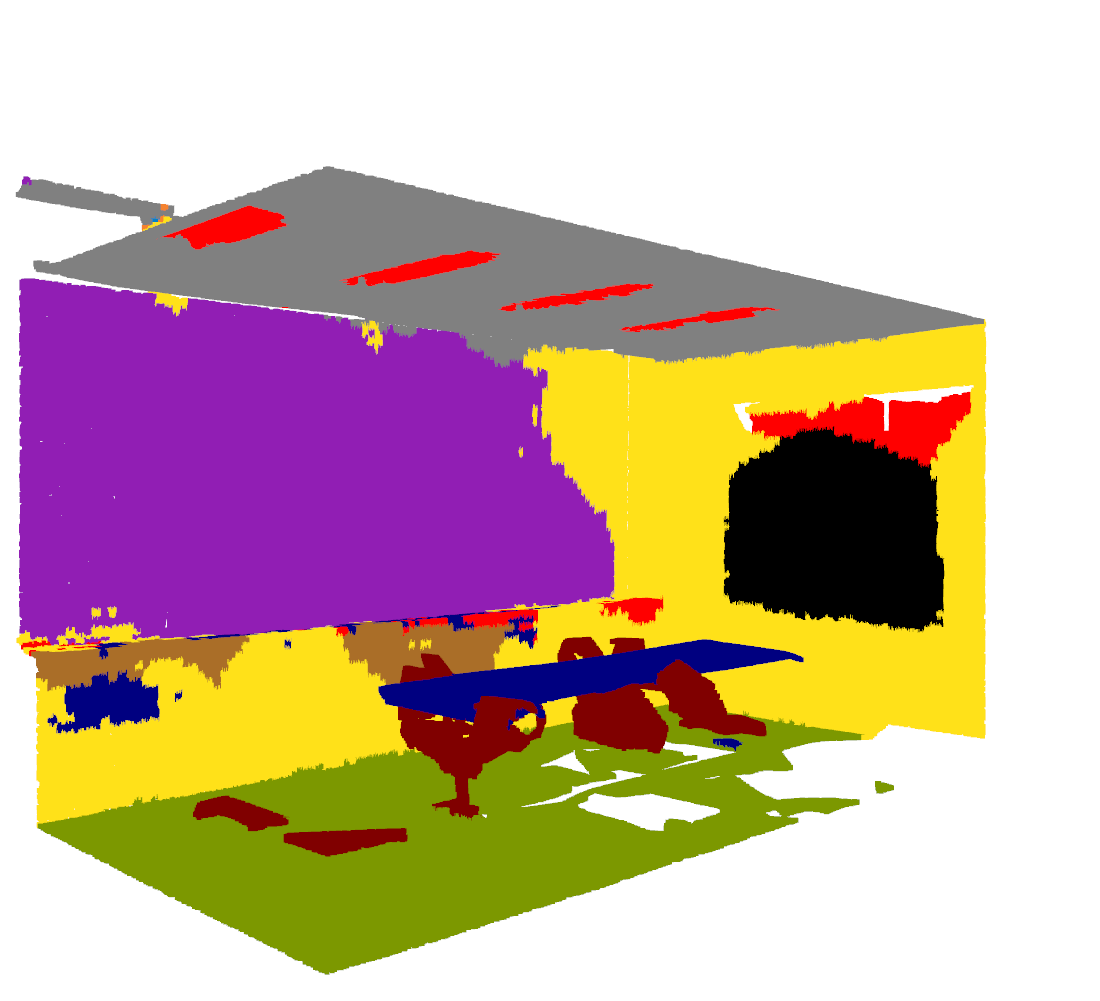}
\includegraphics[width=1\textwidth]{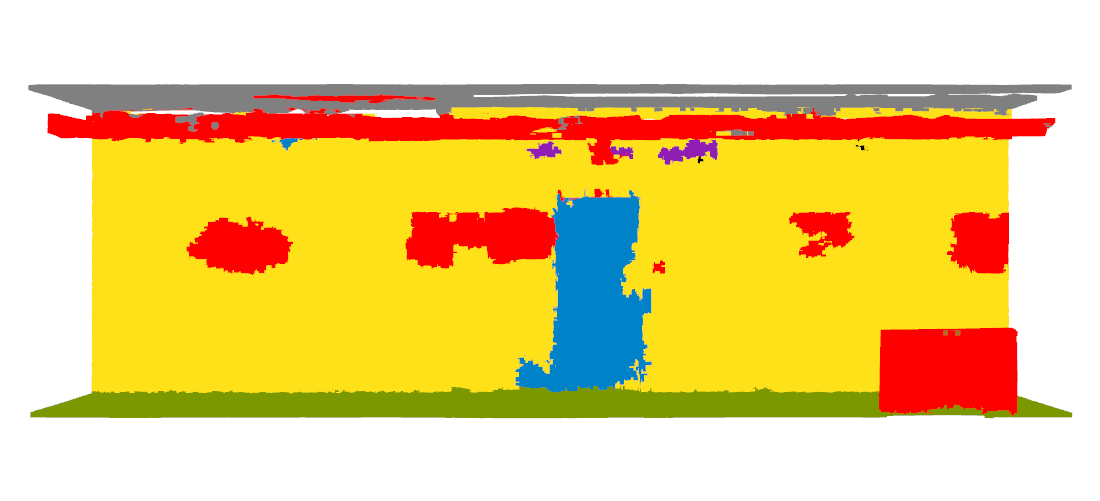}
\includegraphics[width=0.7\textwidth]{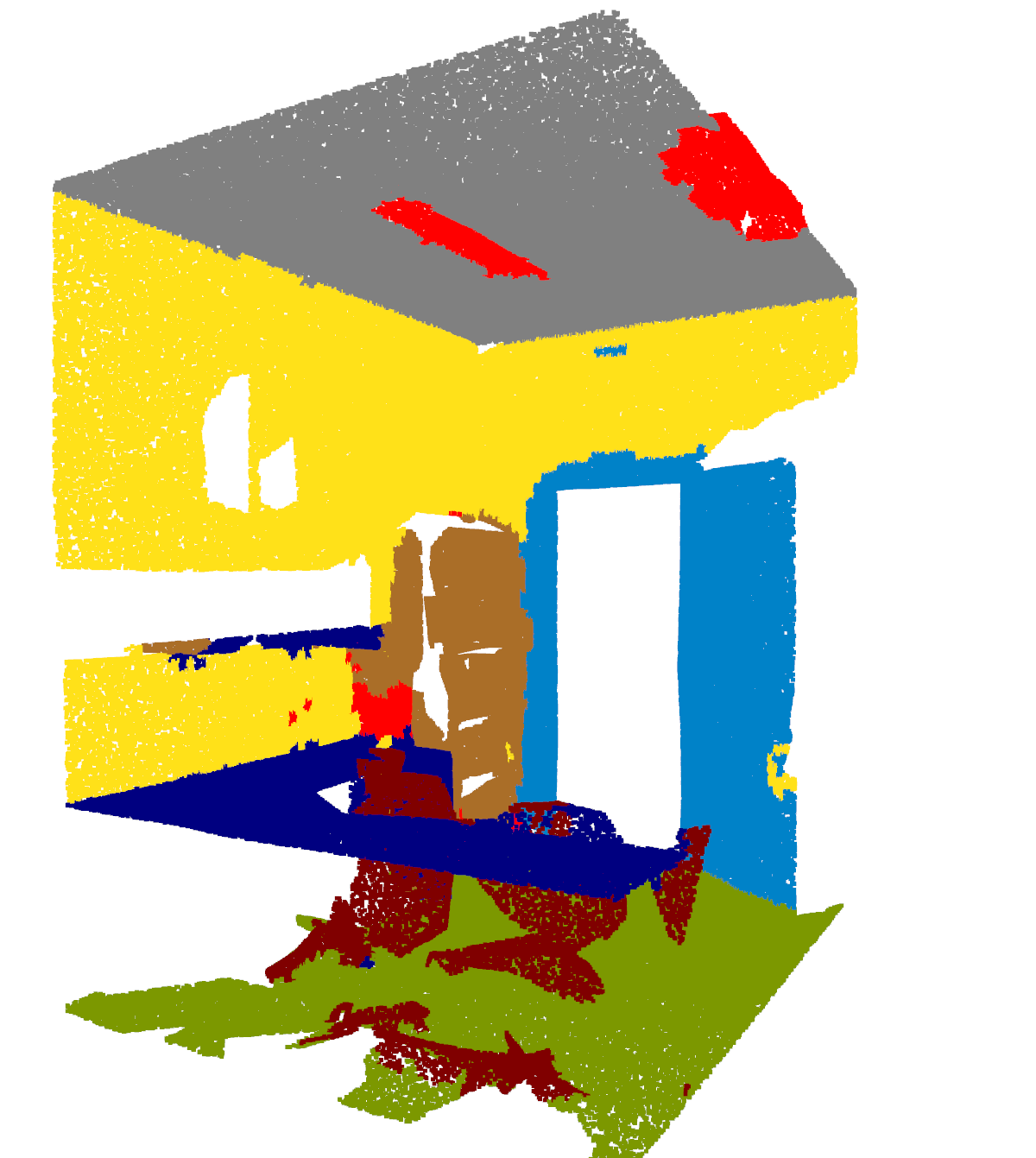}
\end{minipage}
}%
\subfigure[Ground Truth]{
\begin{minipage}[t]{0.24\linewidth}
\centering
\includegraphics[width=1\textwidth]{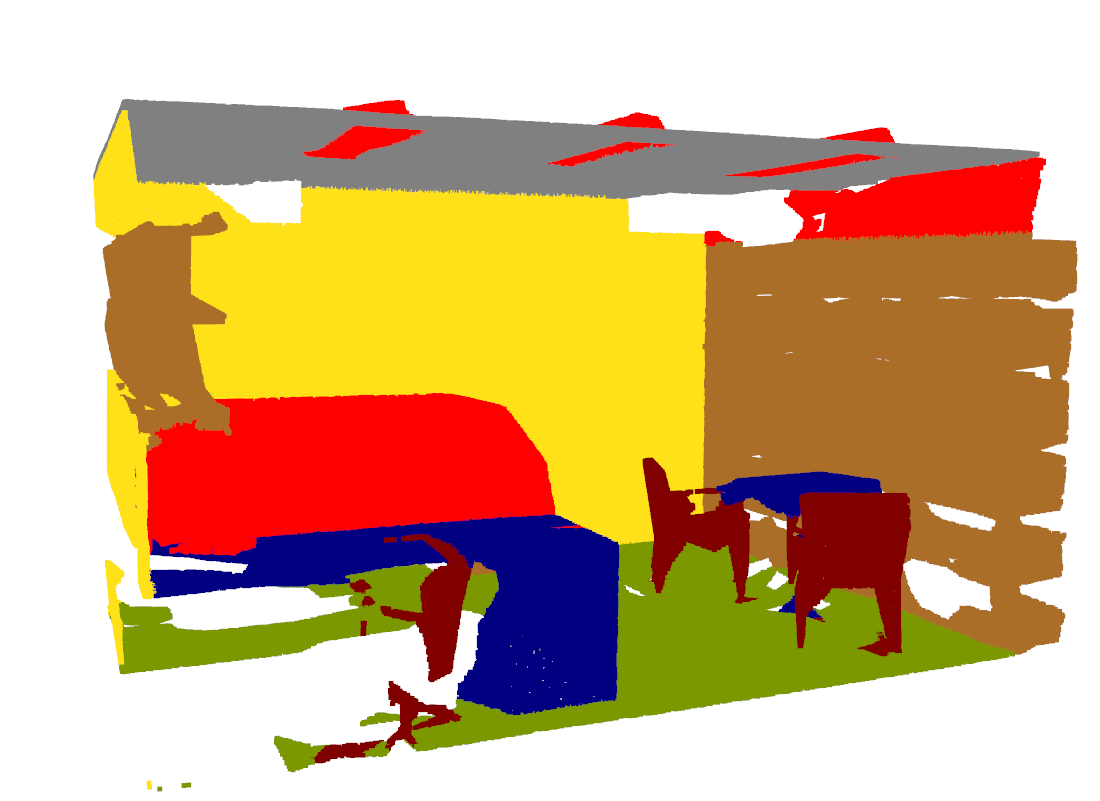}
\includegraphics[width=1\textwidth]{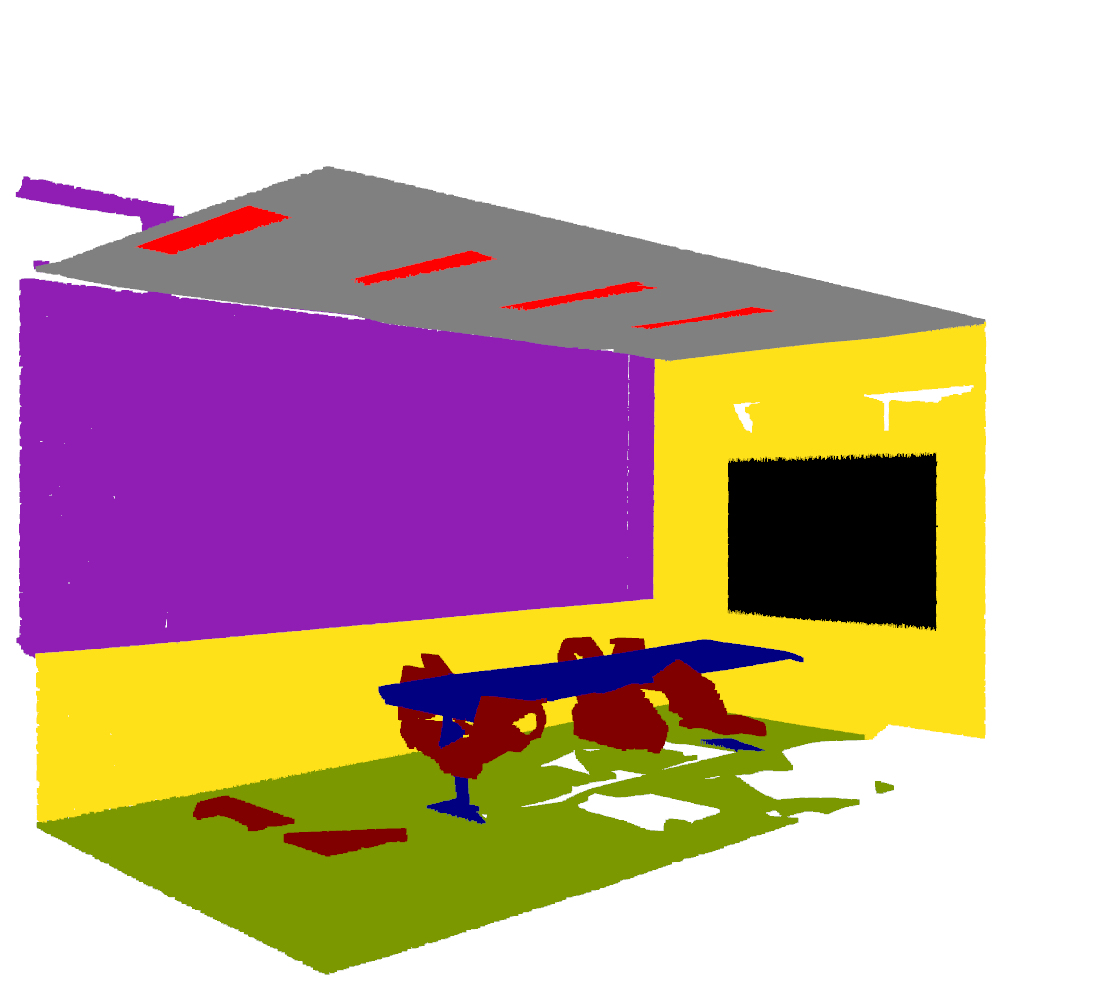}
\includegraphics[width=1\textwidth]{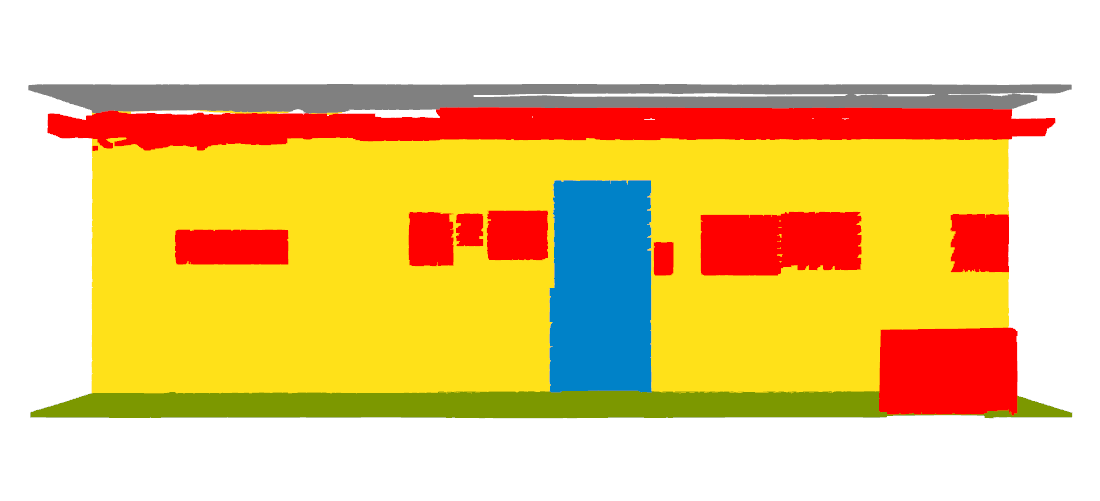}
\includegraphics[width=0.7\textwidth]{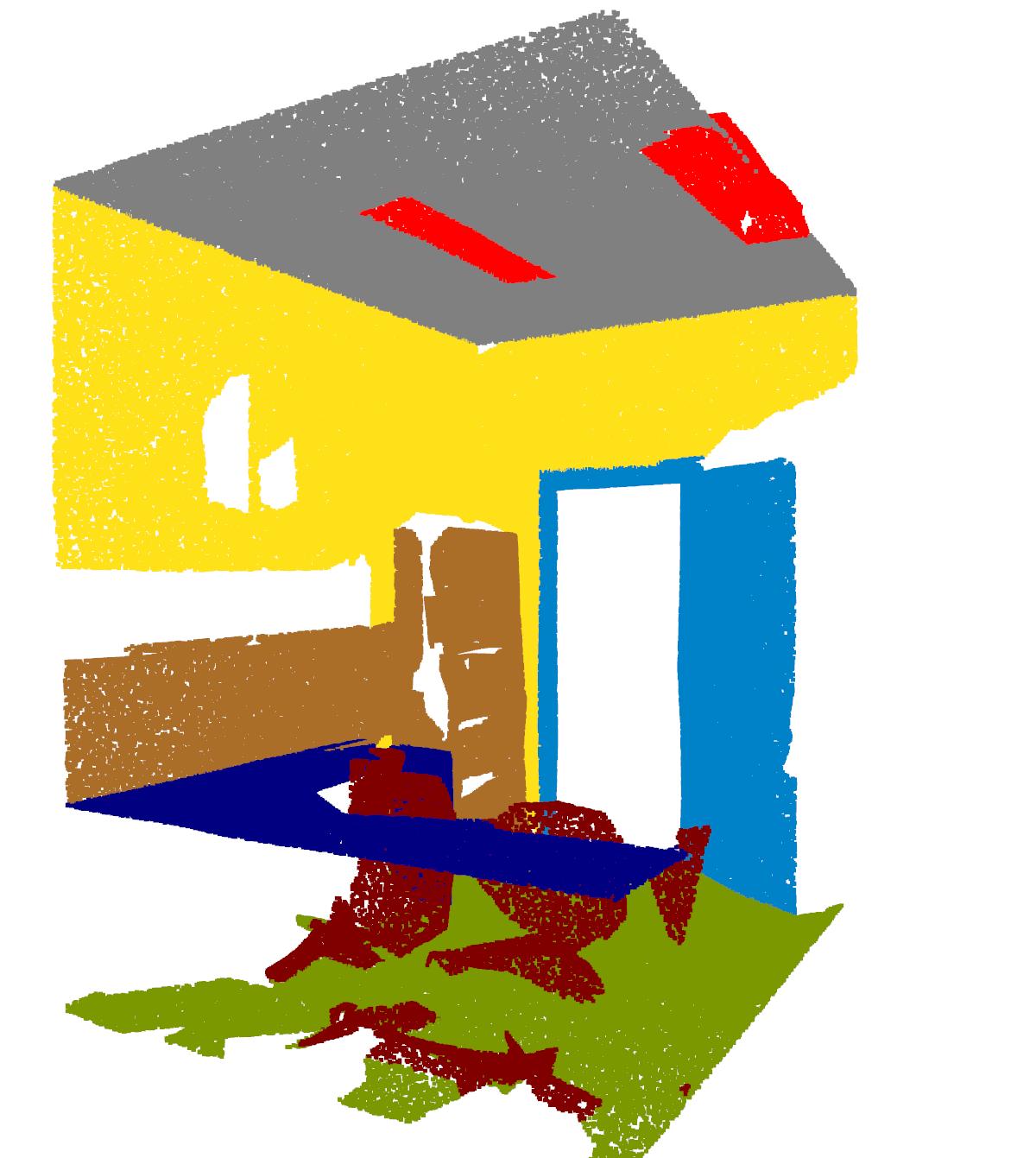}
\end{minipage}
}%
\centering
\caption{Semantic segmentation results on the S3DIS dataset.}
\label{pic:s3dis}
\end{figure*}

\subsection{Main results} \label{Main Results}

Results in this section are obtained under the following settings: receptive field indicator $K=3$, aggregation skeleton is log-activated with $\max\{M\}=256$, and aggregation method is as Eq. (\ref{pooling}).

Segmentation results on the two indoor scene datasets are shown in Table \ref{table1} and \ref{table2}, respectively. We compare our results against both commonly-used benchmarks~\cite{charles_pointnet:_2017,tchapmi_segcloud:_2017,qi_pointnet++:_2017} and state-of-the-art networks that focused on local feature extraction~\cite{tatarchenko_tangent_2018} or global recurrent analysis~\cite{huang_recurrent_2018}. The results show the advantage of using dynamic aggregation to coordinate local and global analysis as a whole.

For evaluation completeness we include some most recent networks~\cite{dai_3dmv:_2018,huang_texturenet:_2019} that provide high performance. However, the input channels fed to these networks are further treated\footnotemark[1].

For the S3DIS dataset, class-wise IoU results demonstrate that our network achieve better prediction on classes that are spatially discrete ([table], [door]) or with rich, compact geometry features ([bookcase], [clutter]), which matches theoretical benefits of forming a self-adaptive pooling skeleton. On the other hand, such performance improvements come with a minor cost on classes that have more uniform geometrical structures (less information density), like [floor] and [wall].

Failure on segmenting class [beam] is natural due to the train-test spilt, as beams in the train-set (Area 1-4, 6) shows a different pattern than that of the test-set (Area 5). We do not use a cross-six-fold validation~\cite{ferrari_3d_2018} to address this matter, as the difference between training data and test data appropriately models real-world applications.

For the ScanNet dataset, class-wise IoU give out similar demonstrations. Discrete objects ([sofa], [bathtub], [toilet], [door]) are better detected whereas structures containing more points yet less information ([wall], [floor]) are partially omitted. We argue that such characteristics are desired, especially for real-world applications like robotic vision or automatic foreground object detection.

Sample segmentation results on the S3DIS dataset are shown in Figure \ref{pic:s3dis}. As T-Conv~\cite{tatarchenko_tangent_2018} did not report class-wise IoU, their results are visualized as benchmarks here for a thorough comparison. As suggested in the text, our network performs particularly well on detecting complicated geometry features and spatial discreteness.

\subsection{Ablation study} \label{ABl}

In this section, we report results from adjusting the dynamic aggregation approach. Unless otherwise specified, all experiments are conducted on the S3DIS dataset.

\textbf{Node receptive field.} The local receptive field size $T_j$, although varying among nodes, can be generally indicated with an average value $g=\sum_jT_j/M=KN/M$. Because $K$ serves as a linear coefficient, its effect is first evaluated, as is shown in Table \ref{table3}. 

\begin{table}[h]
\centering
\scalebox{1}{
\begin{tabular}{c|c|c}
\toprule
$K$ & mIoU & mA \\
\midrule
1 & 54.8 & 63.2\\
2 & 56.1  & 65.3 \\
3 & \textbf{58.8} & \textbf{68.4}\\
4 & 56.2 & 64.2 \\
5 & 55.8 & 63.5\\
7 & 55.1 & 62.9\\
\bottomrule
\end{tabular}}
\caption{\textbf{Reception Field Size: $K$}}
\label{table3}
\end{table}

The global integration network demands a limited skeleton size $M_{max}$ for computational purposes. Fixing $M\equiv M_{max}$ is clearly not desirable. As $N$ varies through two orders in indoor scene datasets~\cite{armeni_3d_2016}, keeping $M$ unchanged leads to a harmful variance on the average receptive field size $g=KN/M$. However, rigidly setting a uniform $g\equiv G$ for all scenes is also not desirable as it fails to take structure complexity into account. An office room and a long hallway may contain the same amount of points, but the former one naturally requires more detailed inspection (Figure \ref{pic:s3dis}).

\begin{figure}[h]
    \centering
    \includegraphics[width=0.4\textwidth]{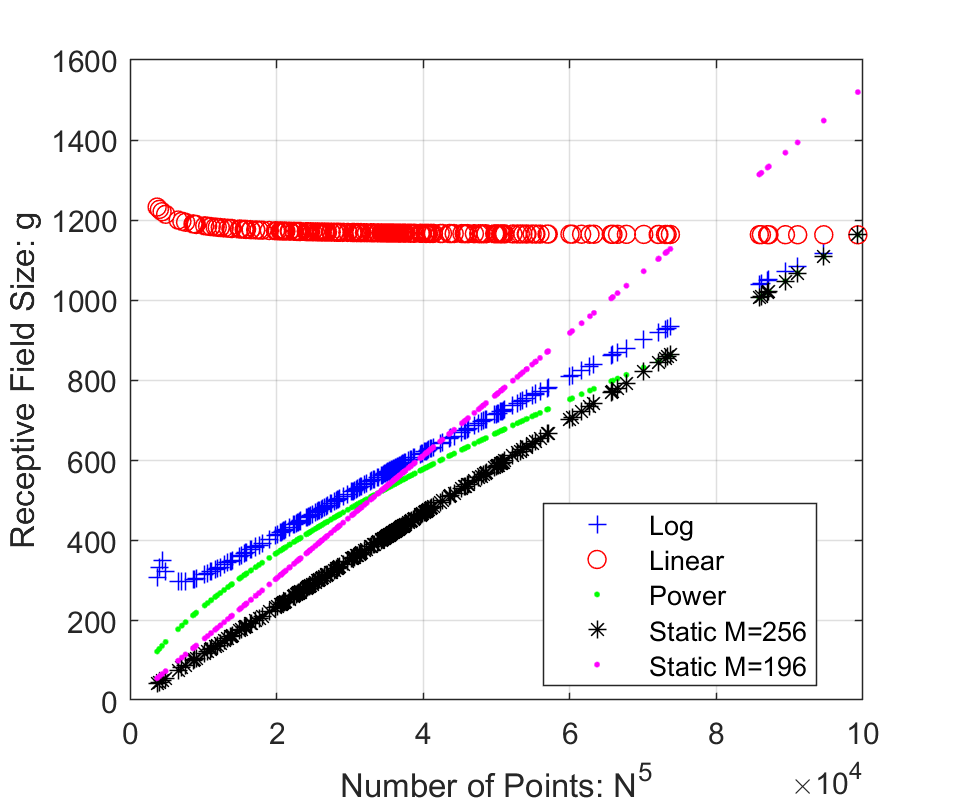}
    \caption{Schematic representation of receptive field size at $K=3$. This figure only represent average trends for demonstration clarity, actual receptive field size varies.}
    \label{fig:rec}
\end{figure}

Without introducing hand-crafted histogram descriptors, a reasonable solution is to adapt a greedy approach: always assigning more nodes (more detailed inspection) than the linear relationship $M=(K/G)\times N$ suggests. Experimental studies show that an approximate logarithm function, $M=-6+70\times\log(N)$, best serves for this purpose as shown in Table \ref{table4}. Still, learning-based approaches can be introduced in the future to explore the parameter space represented by Figure \ref{fig:rec}.

\begin{table}[h]
\centering
\scalebox{1}{
\begin{tabular}{c|c|c}
\toprule
Method & mIoU & mA \\
\midrule
Logarithm & \textbf{58.8} & \textbf{68.4} \\
Power & 57.2 & 66.9 \\
Linear & 56.3 & 63.8 \\
Static (256) & 53.0 & 62.3 \\
Static (196) & 57.4 & 65.6 \\
Static (100) & 45.0 & 53.1 \\
\bottomrule
\end{tabular}}
\caption{\textbf{Receptive Field Size: $M(N)$}}
\label{table4}
\end{table}

\textbf{Aggregation method.} We compare the proposed semi-average aggregation function (Eq. \ref{pooling}) with traditional average/maximum pooling functions. The results shown in Table \ref{table5} indicate advantage from weighting each skeleton node against its receptive field size.

\begin{table}[h]
\centering
\scalebox{1}{
\begin{tabular}{c|c|c}
\toprule
Function & mIoU & mA \\
\midrule
Proposed & \textbf{58.8} & \textbf{68.4} \\
Mean & 57.6 & 64.6 \\
Max & 57.8 & 66.8 \\
\bottomrule
\end{tabular}}
\caption{\textbf{Pooling Function}}
\label{table5}
\end{table}

No significant influence is observed when changing the unpooling method on the S3DIS dataset. However, experiments on the ScanNet dataset suggests otherwise, as is shown in Table \ref{table6}. This phenomenon may be due to scans in this dataset are half-open with less uniform geometrical structures, or that the scale of this dataset being larger and more diverge.

\begin{table}[h!]
\centering
\scalebox{1}{
\begin{tabular}{c|c|c}
\toprule
Function & mIoU & mA \\
\midrule
Mean & \textbf{42.3} & 55.8 \\
Max & 41.1 & \textbf{58.1} \\
\bottomrule
\end{tabular}}
\caption{\textbf{Unpooling Function (ScanNet)}}
\label{table6}
\end{table}

\section{Conclusion}

We present an approach of dynamic aggregation to introduce variance on the extent of multi-level inspection. By introducing self-adapted receptive field size and node weights, dynamic aggregation provides a deep understanding on structures that contain richer geometrical information. We design a network architecture, DAR-Net, to coordinate such intermedium aggregation method with local and global analysis. Experimental results on large-scale scene segmentation indicate DAR-Net outperforms previous network architectures that adopted static feature aggregation.

\bibliographystyle{ieee}
\bibliography{wacv.bib}

\end{document}